\documentclass[11pt]{article}

\usepackage[preprint]{acl}

\usepackage{times}  
\usepackage{helvet}  
\usepackage{courier}  

\usepackage{subcaption}  
\usepackage{natbib}  
\usepackage{caption} 
\usepackage{algorithm}
\usepackage{algorithmic}

\usepackage{amsmath,amsfonts,bm}

\def\eqref#1{equation~\ref{#1}}

\def\1{\bm{1}}

\def\va{{\bm{a}}}
\def\vb{{\bm{b}}}

\def\vr{{\bm{r}}}
\def\vs{{\bm{s}}}

\def\vv{{\bm{v}}}
\def\vw{{\bm{w}}}
\def\vx{{\bm{x}}}

\def\vz{{\bm{z}}}

\def\mA{{\bm{A}}}
\def\mB{{\bm{B}}}

\def\mW{{\bm{W}}}
\def\mX{{\bm{X}}}

\DeclareMathAlphabet{\mathsfit}{\encodingdefault}{\sfdefault}{m}{sl}
\SetMathAlphabet{\mathsfit}{bold}{\encodingdefault}{\sfdefault}{bx}{n}

\def\gD{{\mathcal{D}}}

\def\gG{{\mathcal{G}}}

\def\gL{{\mathcal{L}}}
\def\gM{{\mathcal{M}}}

\def\gO{{\mathcal{O}}}
\def\gP{{\mathcal{P}}}

\def\gR{{\mathcal{R}}}

\usepackage{url}
\usepackage{verbatim}
\usepackage{amsthm}
\usepackage{multirow}
\usepackage{booktabs}
\usepackage{array}
\usepackage{amssymb}

\allowdisplaybreaks[2]
\def\ie{\textit{i.e.,~}}

\def\1{\mathbbm{1}}

\usepackage{times}
\usepackage{latexsym}

\usepackage[T1]{fontenc}

\usepackage[utf8]{inputenc}

\usepackage{microtype}

\usepackage{inconsolata}

\usepackage{graphicx}
\theoremstyle{plain}
\newtheorem{theorem}{Theorem}[section]

\theoremstyle{definition}

\newtheorem{conjecture}[theorem]{Conjecture}
\theoremstyle{remark}

\title{Tackling the Inherent Difficulty of Noise Filtering in RAG}

\author{Jingyu~Liu\textsuperscript{1},Jiaen~Lin\textsuperscript{3},Yong~Liu\textsuperscript{1,2}\thanks{Corresponding Author.} \\
	\textsuperscript{1}Gaoling School of Artificial Intelligence, Renmin University of China, Beijing, China\\
	\textsuperscript{2}Beijing Key Laboratory of Big Data Management and Analysis Methods, Beijing, China \\
	\textsuperscript{3}School of Software Tsinghua University, Beijing, China\\
	\texttt{liujy1016@ruc.edu.cn} \\
}

\begin{document}
\maketitle

\begin{abstract}
Retrieval-Augmented Generation (RAG) has become a widely adopted approach to enhance Large Language Models (LLMs) by incorporating external knowledge and reducing hallucinations. However, noisy or irrelevant documents are often introduced during RAG, potentially degrading performance and even causing hallucinated outputs. While various methods have been proposed to filter out such noise, we argue that identifying irrelevant information from retrieved content is inherently difficult and limited number of transformer layers can hardly solve this. Consequently, retrievers fail to filter out irrelevant documents entirely. Therefore, LLMs must be robust against such noise, but we demonstrate that standard fine-tuning approaches are often ineffective in enabling the model to selectively utilize relevant information while ignoring irrelevant content due to the structural constraints of attention patterns. To address this, we propose a novel fine-tuning method designed to enhance the model's ability to distinguish between relevant and irrelevant information within retrieved documents. Extensive experiments across multiple benchmarks show that our approach significantly improves the robustness and performance of LLMs.
\end{abstract}
\section{Introduction}

Large Language Models (LLMs) \citep{GPT} have demonstrated remarkable capabilities across a variety of tasks, including text generation and question answering \citep{LLM_powerful,CoT}, code generation \citep{llm_code}, and information retrieval \citep{llm_IR}. However, current LLMs often suffer from serious hallucinations \citep{LLM_hallucination,RAG_defense_attack_25} due to a lack of factual information. Moreover, the knowledge embedded within LLMs is encoded in their parameters \citep{memory3}, meaning that incorporating new knowledge requires further fine-tuning, which is both time-consuming and resource-intensive. Consequently, augmenting LLMs with external retrievers has led to significant performance improvements \citep{RAG,KAG, RAG_survey, RAG_JMLR,RAG_noise_finetune_25}.

However, in real-world RAG scenarios, the information retrieved from documents is not always directly usable and often requires further processing because documents may contain noisy information \citep{longllmlingua, llmlingua}, and some documents may even be completely distracting, containing incorrect answers \citep{RAG_distraction, RAG_noise_skew,RAG_noise_finetune_25,RAG_robust_finetune_25}. Such noise and distracting documents can negatively impact performance.

Apparently, to improve the performance, we can either reduce the number of distracting documents by more advanced retriever \citep{RAG_duality,RAG_robust,RAG_corrective} or fine-tune the model \citep{RAG_noise_finetune_25,RAG_robust_finetune_25,RAG_robust} to make it more robust to noisy information. Our paper shows that these two methods fail because,

\begin{itemize}
    \item Filtering out irrelevant information is inherently difficult, small retrieval models fail to solve it.
    \item Fine-tuning the LLM can hardly distinguish irrelevant information while taking advantage of relevant ones.
\end{itemize}
 
About the difficulty of filtering out irrelevant information.
Considering the query \textit{`Alice and Bob are running, Bob is exhausted. How does Bob feel?'} Here, assessing the relevance of \textit{`exhausted'}  necessitates considering the token \textit{`Bob'}, \textit{`feel'} in the query alongside \textit{`Bob'} which is the subject of \textit{`exhausted'}. Thus, evaluating relevance requires the information from three or even more tokens. Yet, the attention mechanism typically computes only pairwise relationships, making it challenging to resolve this issue within a limited number of transformer layers \citep{3Match}.
Therefore filtering out noise documents is inherently difficult which explains why current small retriever models \citep{Contriever,BM25,DPR} would always incorporate some noisy information in the retrieval results.

Therefore, a question is can powerful LLMs solve this problem. We hope that the LLM could be robust to noisy information while extracting helpful information \citep{RAG_robust,RAG_robust_finetune_25}.
However, we argue that standard fine-tuning is structurally ill-suited for this task. The core issue lies in a fundamental trade-off imposed by their linear update mechanism. To filter noise, the learned parameter update must apply a strong negative adjustment to the attention scores of irrelevant tokens. Yet, this same linear adjustment is applied across all tokens, which can inadvertently distort the nuanced, relative attention patterns among the relevant tokens that are crucial for complex reasoning. In essence, the model is forced to choose between effective noise filtering and preserving its reasoning capacity.

To overcome this limitation, our work decouples these competing objectives. We propose a novel fine-tuning method that introduces a non-linear rectification function to the attention update. This function is specifically designed to operate in two distinct regimes: for irrelevant tokens, it creates a sharp, saturating penalty to effectively "zero them out"; for relevant tokens, it allows for more gentle adjustments that preserve their relative importance. This approach enables the model to aggressively filter noise while simultaneously safeguarding its core reasoning abilities. Extensive experiments show that our method significantly improves robustness in noisy RAG settings.




The main contributions of this paper are:
\begin{itemize}
    \item We reveal that the inherent triple-wise nature of the noise filtering process may necessitate numerous transformer layers, which means we can hardly filter out noise documents with small models.
    \item We show that fine-tuning the LLM to filter out noise while effectively taking advantage of the relevant information is challenging
    \item We developed a new fine-tuning method to better distinguish irrelevant tokens and extensive experiments shows its effectiveness.
\end{itemize}

\section{Related Work}

  \textbf{RAG in LLM.} Many recent works have explored better RAG strategies. \citet{RAG_replug} treat the model as a black box and design a retriever to enhance performance. 
  However, researchers have identified that noise in the context can negatively impact performance \citep{RAG_distraction,retrieval_noise_25_1,RAG_robust_finetune_25}, some researchers focus on eliminating this noisy information and compress the noisy documents\citep{longllmlingua,long_context_llm,long_context_medical,prompt_compression}. 
  And some others try to fine-tune the large language model to make it more robust to noisy information\citep{RAG_robust,RAG_RAFT,RAG_noise_finetune_25}. 
  
  \textbf{Filtering the noise.} \citet{RAG_robust} tries to fine-tune the LLM to better filter out distracting documents, while RAFT \citep{RAG_RAFT} uses different proportion of distracting documents to make it more robust. Self-RAG\citep{self_RAG}, use special tokens to indicate whether the LLM should use the document information.. \citet{RAG_duality} first highlighted the duality of RAG, encompassing both benefits and detriments. Also, various researchers propose new fine-tuning methods to make the LLM robust to noisy information \citep{RAG_filter1,RAG_robust_finetune_25,RAG_noise_finetune_25}. Also \citet{RAG_robust_finetune_25} finds that some specifically designed fine-tuning methods helps less in modern models. But they fail to understand why noise filtering is difficult in current LLMs. Differential Transformer \citep{differentialtransformer} calculates the attention score as difference between two separate softmax attention to cancel noise. Also CrAM \citep{cram} tries to adjust attention weights given the credibility of different documents.

\section{The Triple-Wise Problem}

Clearly, noise in the documents adversely affects the performance of large language models (LLMs). And various researchers are trying to develop new retrieval methods to reduce the number of noisy information, and some others try to develop a filtering model to filter out the irrelevant information before input to LLM. In this section, we show that filtering out noise documents is inherently a complex problem. It is difficult to filter out noise by small model based rerankers/retrievers.

\begin{figure}
    \centering
    \includegraphics[width=1\linewidth]{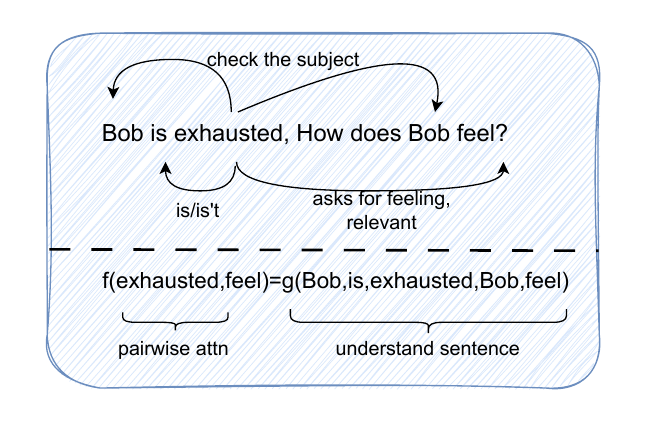}
    \caption{Judging the relevance of the token 'exhausted' actually requires checking the whole meaning of the sentence, so it can hardly be done by limited number of transformer layers.}
    \label{fig:triple_wise_problem}
\end{figure}

For input $\mX = [\vx_0^T, \vx_1^T, \ldots, \vx_{n-1}^T]$ with the query and document, and we should decide is the document relevant to query. It's crucial to note that calculating the relevance of the document require the involvement of many tokens. For instance, in the query \textit{“Alice and Bob are running, Bob is exhausted. How does Bob feel?”}, the context \textit{“exhausted”} describes the feeling of Bob, which serves as a useful contenxt. However, determining relevance necessitates considering \textit{“Bob”}, \textit{“feel”} in the query alongside \textit{“Bob”}, which is the subject of \textit{“exhausted”}, only in this way the model can understand that “exhausted” describes Bob and the query asks for the feeling of Bob. We show it more clearly in Figure \ref{fig:triple_wise_problem}. Therefore, identifying the relevance of a token demands information from multiple tokens.
However, self-attention computes relationships only between pairs, making it challenging to effectively address this issue.

As self attention only calculates pair-wise relationship between tokens, and judging the relevance of the document requires the involvement of multiple tokens. It is necessary to stack multiple attention layers to aggregate information from different tokens to conduct judgment.
However, as noted in \citet{3Match}, To effectively solve the triple-wise problem, we need the multi-layer transformer to have width, depth, embedding dimension, or bit complexity at least $N^{\Omega(1)} c$, where $N$ is context length and $c$ is a constant represents the embedding dimension required to represent noise filtering information. This is impractical for small models as it only shows limited depth, width and embedding dimension.

The challenge arises from the triple-wise nature of the problem contrasted with the pairwise nature of attention; the model can only evaluate a token's relevance when its embedding contains substantial information about the whole context. 
For example, as shown in Figure \ref{fig:triple_wise_problem}, to assess the token \textit{“exhausted”} in the sentence \textit{“Bob is exhausted. How does Bob feel?”}, the embedding must encompass information about its subject of \textit{“exhausted”} and the subject of \textit{“feel”}, \textit{“Bob”}, as well as contextual details from phrases like \textit{“is/isn't”} and \textit{“feeling”}. Therefore, a significant amount of information must be incorporated into the embedding before any judgment can be made, it actually requires the understanding about the meaning of the whole sentence before judging the relevance of the token.

However, a single layer of self-attention can only consider the input of fixed dimension embedding, which contains information up to $mp$, where $m$ represents the embedding dimension and $p$ indicates precision of embedding. The term $mp$ signifies the maximal information carried by the embedding. But this can hardly represent the meaning of the whole sentence, especially when the input sequence is long. 


This indicates that trying to filter out irrelevant context is difficult, the input to LLM would like to incorporate noisy information in the retrieved documents. However, current large language models holds great embedding dimension (4096 for Llama3-8B) and depth (32 for Llama3-8B), which should be able to solve the triple-wise problem theoretically.
This suggests the burden of noise filtering should shift from the limited-capacity retriever to the powerful LLM itself. However, as we will show next, making the LLM robust is not a straightforward task.

\section{Robustness of LLM When Faced with Noise}





Clearly, noise in the documents negatively impacts the performance of LLMs, and noisy information is inevitable in RAG. Therefore some researchers focus on fine-tuning the model to make it more robust when faced with noisy information thereby enhancing performance \citep{RAG_robust, RAG_noise_finetune_25,RAG_robust_finetune_25, longllmlingua}. And there is a question that, can the fine-tuned LLM effectively filter out irrelevant information and gather useful information to get the final answer?


Let $\vr$ represents the relevance of tokens, $r_i=0$ means that token $\vx_i$ is a noise token, otherwise the token is relevant. Let $attn(\vx_i, \vx_j) = (\mW_q \vx_i)^T \mW_k \vx_j$ represent the attention pattern which is trained to extract relevant information.

And we want to filter out irrelevant information while preserving the attention pattern on relevant tokens, this can be represented as following:

\begin{equation}
  \sigma\left(\widehat{attn}(\vx_i,\vx_j)\right)=
  \begin{cases}
    0 & \text{ if } \ r_j=0, \\
    \sigma\left((\mW_q \vx_i)^T \mW_k \vx_j \right) & \text{ else, }\\
  \end{cases}
  \nonumber
\end{equation}

where $\sigma$ means the softmax function. This shows that the desired attention pattern should effectively exclude noise while preserving the attention pattern of relevant tokens.

Therefore, $\widehat{attn}$ can be considered the optimal response when confronted with noise, as it effectively filters out irrelevant tokens and utilizes the relevant information in the most efficient manner.

Fine-tuning the model involves adjusting its parameters and attention pattern, which allows the fine-tuned model to be expressed as 

\begin{equation}
    \begin{split}
        attn'(\vx_i, \vx_j) &= \left((\mW_q + \Delta \mW_q) \vx_i\right)^T  \left(\mW_k + \Delta \mW_k\right) \vx_j\\
        &= \vx_i^T (\mW + \Delta \mW) \vx_j,
    \end{split}
    \nonumber
\end{equation}

where $\mW = \mW_q^T \mW_k$ and $\Delta \mW$ represents the adjustments. The critical question arises: can we fine-tune the model to approximate the optimal one, i.e., is there a $\Delta \mW$ such that $\sigma(attn'(\vx_i, \vx_j)) \approx \sigma(\widehat{attn}(\vx_i, \vx_j))$? $\sigma(\cdot)$ represents the softmax.

\begin{theorem} \label{theorem:lora_fail}
  if there exists $$attn'(\vx_i,\vx_j)=\vx_i (\mW+\Delta \mW) \vx_j,$$ $\epsilon$ approximates $\widehat{attn}(\vx_i,\vx_j)$ \ie 
  \begin{equation}
        1-\epsilon \leq \frac{\sigma(attn'(x_i,x_j))}{\sigma(\widehat{attn}(x_i,x_j))} \leq 1+\epsilon,
      \nonumber
  \end{equation}
  where $\sigma$ represents softmax, then we need
  \begin{equation}
    \begin{split}
      \xi_{r} \lesssim \ln \frac{1}{1-\epsilon},
    \end{split}
    \nonumber
  \end{equation}
  where $\xi_{r}=\max(\vx_i^T \Delta \mW \vx_j)-\min (\vx_i^T \Delta \mW \vx_k),$ and $x_{j,k}$ is relevant tokens.
\end{theorem}

Apparently, if we want to approximate the optimal reasoning pattern, we need $\xi_r \approx 0$, so $\xi_{r} \lesssim \ln \frac{1}{1-\epsilon} \approx 0$.
This shows that, to effectively fine-tune the model to filter out irrelevant information in the attention matrix, we need $\xi_{r} \approx 0$. This implies that for all tokens to be retained, $\vx_i^T \Delta \mW \vx_j$ must remain nearly constant. A solution could make $||\Delta \mW||$ small or even $0$, but this actually also requires to split the tokens to be retained and to be filtered, so a small $||\Delta \mW||$ does not solve the problem as shown in Appendix \ref{proof_lora_fail}. As a result, approximating the optimal solution proves to be quite challenging because the attention will be disturbed due to the noise filtering fine-tuning. 


One might argue that preserving the original attention pattern is unnecessary. Perhaps the model could learn a new, superior attention pattern during fine-tuning that excels at both filtering noise and leveraging semantic information. However, this perspective overlooks a fundamental conflict between these two objectives.

To conduct the two task, we require the input to attention contain two different kinds of information, noise filtering and semantic reasoning.
As suggested by \citet{IB_generalization}, a model with finite capacity attempts to optimize for two different tasks would show suboptimal performance on both task because the two different tasks requires different information, and each act as noise to each other, so when the model is trying to filter out noise, but some semantic information is also incorporated into the input, then it will downweight the performance on the filtering process and vice versa.

This also fits for the feed forward layers. It cannot be an expert filter and an expert reasoner at the same time when its input is already contaminated. Therefore, relying on the FFN to clean up the mess made by the attention layer is an inefficient strategy that compromises the model's overall performance. We show more analysis in Appendix \ref{proof_mlp_fail}

It is worth noting that although the analysis mainly focus on single head attention, for multi head attention, we can use one head focus on filtering out noise and another head focus on taking advantage of the relevant information. This is indeed ideal, but fine-tuning based on existing LLMs fails to greatly influence the existing parameters, otherwise it will cause catastrophic forgetting \citep{catastrophic_forgetting1,catastrophic_forgetting2}.

\section{Fine-tuning for Noise Filtering}

\subsection{Attention Rectification}

Conventional fine-tuning paradigms, primarily adjust the model's behavior by introducing an update matrix, $\Delta \mW$, to the original weight matrix $\mW$. The modified attention score between a query token $i$ and a key token $j$ is computed based on their embeddings, $\vx_i$ and $\vx_j$. However, as noted in Theorem \ref{theorem:lora_fail}, a simple linear addition of the update term $\vx_i^T \Delta \mW \vx_j$ faces a fundamental trade-off: it struggles to simultaneously (1) create a sufficiently large margin between the attention scores of relevant and irrelevant tokens, and (2) preserve and optimize the nuanced, relative attention patterns among multiple relevant tokens, which is crucial for the model's intrinsic reasoning capabilities.

\begin{figure}
    \centering
    \includegraphics[width=0.9\linewidth]{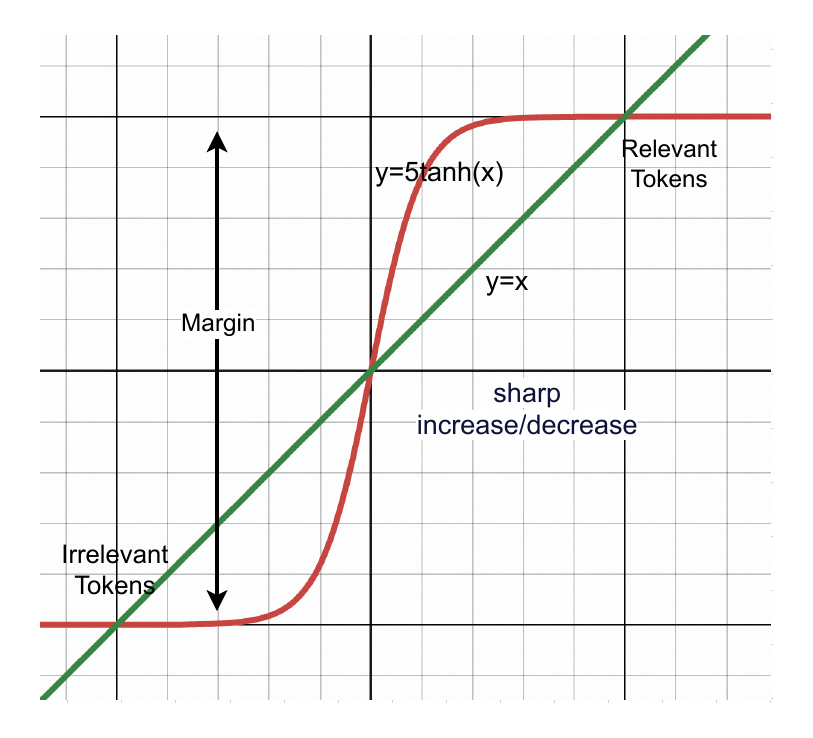}
    \caption{plot of $5\text{tanh}(x)$, this shows that, by tanh, we can effectively enlarge the margin between relevant and irrelevant tokens and maintain similar attention weight to those relevant ones.}
    \label{fig:tanh}
\end{figure}

To address this, we propose to augment the standard attention mechanism with a non-linear rectification function, $g(·)$, applied to the attention update. The final attention is computed as:
\begin{equation}
attn'(\vx_i, \vx_j) = \vx_i^T \mW \vx_j + g(\vx_i^T \Delta \mW \vx_j).
\label{eq:dar_main}
\end{equation}
The core of our method lies in the design of the rectification function g(x). The function g(x) is designed to operate in two distinct regimes: a filtering regime for low-relevance tokens and a refinement regime for high-relevance tokens. This is achieved by the following formulation:

\begin{equation}
  g(x)=
  \begin{cases}
    \max(\xi \cdot \tanh(x), x) & \text{ if } \ x>=0, \\
    \min(\xi \cdot \tanh(x), x) & \text{ else, }
  \end{cases}
  \nonumber
\end{equation}

where $\xi$ is a hyperparameter that controls the saturation threshold. The behavior of $g(x)$ is illustrated in Figure \ref{fig:tanh}.
For small or negative values of the attention update $x$, the term $\xi * tanh(x)$ dominates. The hyperbolic tangent function offers two key advantages:

\begin{itemize}
    \item Sharp Discrimination: The steep gradient of $tanh(x)$ around $x=0$ creates a sharp transition, effectively amplifying the margin between positive (potentially relevant) and negative (irrelevant) attention updates. This serves as a powerful mechanism for filtering out noise as a large negative attention could be allocated to unrelated tokens.
    \item Saturating Behavior: As $x$ becomes sufficiently large, $tanh(x)$ approaches 1, causing the output to saturate at $\xi$. This ensures that all highly relevant tokens receive a consistent and significant attention boost, preventing their original relative attention scores (after Softmax) from being severely distorted. The scaling factor $\xi$ is chosen to be large enough to establish a clear margin, determined by the typical attention score variance in the base model.
\end{itemize}

In this way, we can effectively separate the relevant and irrelevant tokens during the sharp discrimination, and we can ensure that the relevant tokens share similar attention score.

However, simply using the $\xi \cdot tanh(x)$ may not be enough, a key limitation of using $\xi \cdot tanh(x)$ alone is that it clamps the attention boosts for all highly relevant tokens to a single value $\xi$, precluding any further fine-grained adjustments among them. Our composite function $g(x)$ overcomes this. When the attention update $x$ is large enough to exceed $\xi \cdot tanh(x)$, $g(x)$ reverts to a linear identity which means $g(x) = x$. This linear growth phase allows the model to continue differentiating among highly relevant tokens, enabling it to learn a more optimal attention distribution for the specific downstream task, rather than merely preserving the original one.

From a regularization standpoint. We actually require the attention module to conduct two tasks, nose filtering and relevant information aggregation. The two tasks may require different information to process and they are actually noise to each other. In this way, an unconstrained linear update might learn to assign attention with high variance scores based on such noise as shown in \citet{IB_generalization}. 
The saturating nature of the $tanh$ component in $g(x)$ acts as a "soft clamp" constraining these potentially high variance values within a bounded range ($\xi$). This restriction prevents the model from becoming overly sensitive to noisy features, thereby enhancing its robustness and generalization performance. For instance, an attention update that might vary between $\xi \pm \delta$ with only linear update, and with the tanh activation, it would be regularized to $\xi \pm \epsilon$, where $\epsilon << \delta$.

Also, $\max(\cdot)$ and $\min(\cdot)$ is not a continuous function, so we use 
\begin{equation}
\begin{split}
    g(x)
  &=\begin{cases}
    \max(\xi \cdot \tanh(x), x) & \text{ if } \ x\geq 0, \\
    \min(\xi \cdot \tanh(x), x) & \text{ else, }
  \end{cases} \\
    &\approx \log(\exp(a)+\exp(b)+1)\\
    &\quad -\log(\exp(-a)+\exp(-b)+1),
\end{split}
\nonumber
\end{equation}
where $a=\xi \cdot tanh(x)$, $b=x$. Apparently, when $a,b>=0$, $g(x)$ will be dominated by $\log(\exp(a)+\exp(b)+1)$, which is approximately $max(a,b)$. Otherwise if $a,b<0$, $g(x)$ will be dominated by $-\log(\exp(-a)+\exp(-b)+1)$, which is approximately $min(a,b)$.

\subsection{The Auto-Regressive Nature may Cause Problem}
\label{section_simplify}

With the activation, the attention module can effectively filter out noise information.
However, most LLMs are trained in an auto-regressive manner, meaning that a token cannot aggregate information from any tokens that appear later in the sequence; thus, $a_{i,j} = 0$ if $j > i$. Typically, the query is positioned after the document tokens, preventing these document tokens from assessing their relevance effectively because they have no access to the query. Consequently, the relevance judgment must occur during the calculation of the query embeddings, which is usually much shorter than the document, making the nose filtering harder because we need to judge the relevance of $n_{doc}$ tokens during the calculation of $n_{query}$ tokens.

Instead, if we position the query ahead of documents, then the relevance can be effectively calculated during the calculation document token embedding. This arrangement enables the information of query to be effectively transferred to the document tokens for relevance judgment. In this way we can judge the relevance of $n_{doc}$ tokens during the calculation of $n_{doc}$ tokens, as $n_{doc}$ is usually much larger than $n_{query}$, so placing the query ahead could help. Therefore, we can hypothesize that when there is noise in the document, placing the query at the beginning would help the judgment. We conduct experiments on various datasets in Appendix \ref{app:more_experiments} to show that placing the query ahead can effectively help the performance.

\begin{table*}[htbp]
  \centering
\resizebox{\textwidth}{!}{%
    \begin{tabular}{cccccccccccc}
    \toprule
          & \multicolumn{2}{c}{NQ} & \multicolumn{2}{c}{TriviaQA} & \multicolumn{2}{c}{HotpotQA} & \multicolumn{2}{c}{2wiki} & \multicolumn{2}{c}{ASQA} &  \\
          & reverse & vanilla & reverse & vanilla & reverse & vanilla & reverse & vanilla & reverse & vanilla & mean \\
    \midrule
    vanilla & 52.4  & 46.7  & 52.1  & 45.6  & 61.4  & 54.3  & 53.7  & 52.1  & 42.0  & 41.8  & 50.2  \\
    LoRA  & 65.7  & 62.6  & 72.6  & 71.3  & 86.1  & 85.6  & 95.3  & 96.4  & 42.3  & 42.2  & 72.0  \\
    $\xi=1$     & 64.9  & 64.4  & 73.9  & 73.1  & 86.6  & 86.1  & 95.6  & 96.5  & 44.3  & 44.3  & 73.0  \\
    $\xi=3$    & 66.7 & 64.2  & 74.1  & 73.1  & \textbf{87.3 } & \textbf{86.3 } & 97.6  & 97.2  & \textbf{47.8 } & \textbf{46.4 } & 74.1  \\
    $\xi=5$     & \textbf{67.6}  & 64.8  & \textbf{74.5 } & \textbf{73.7 } & 87.1  & 86.1  & 97.3  & \textbf{97.5 } & 47.2  & 46.1  & 74.2  \\
    $\xi=10$    & 66.9  & \textbf{65.4 } & 74.1  & 73.4  & 87.2  & 85.8  & \textbf{97.9 } & 97.4  & 47.3  & 46.2  & \textbf{74.2 } \\
    \bottomrule
    \end{tabular}%
    }
  \caption{Performance of our fine-tuning method when faced with explicit distracting documents. reverse means we place the query ahead of documents and vanilla means the query is placed after documents}
  \label{tab:per_noise}%
\end{table*}%

\section{Experiments} \label{sec:experiments}

\subsection{Datasets and Metrics}
To evaluate the performance of our proposed method, we use Nature Questions \citep{natural_question}, TriviaQA \citep{TriviaQA} which is traditionally used to evaluate the noise robustness of RAG system. Also, we use multi hop reasoning datasets HotpotQA \citep{hotpotqa} and 2WikiMultiHopQA \citep{2wikimultihop} as well as the long form QA dataset ASQA \citep{ASQA} to show the performance of our method. For the first 4 datasets, we use accuracy to measure the performance which is determined by whether the predicted answer contains the ground-truth answer. For ASQA, we measure the percentage of short answers are shown in the generated answer to evaluate the performance. 

For all 5 datasets, we use Dense Passage Retriever \citep{DPR} as the retriever, we retrieve some documents and select the first 3 documents that are not presented in the gold documents as the noisy documents, then combine the noisy documents with the gold documents as the input to the LLM, showing that our method can effectively distinguish distracting documents while taking advantage of relevant ones.

For the first 4 datasets, we randomly select 3000 samples to test and another 7000 to train. For ASQA we use the split of ALCE \citep{ALCE} and use the 948 samples to test the performance and another 4000 for training.
More Experimental settings can be seen in Appendix \ref{app:more_experiments}

\begin{table}[htbp]
  \centering

  \resizebox{0.48 \textwidth}{!}{%
    \begin{tabular}{ccccccc}
    \toprule
          &       & NQ    & Trivia & Hotpot & 2wiki & ASQA \\
    \midrule
    \multirow{3}[2]{*}{Qwen 7B} & vanilla & 53.5  & 59.9  & 74.8  & 67.1  & 43.5  \\
          & LoRA  & 61.4  & 69.2  & 82.3  & 95.2  & 46.5  \\
          & tanh  & \textbf{62.7 } & \textbf{69.9 } & \textbf{84.3 } & \textbf{96.3 } & \textbf{48.3 } \\
    \midrule
    \multirow{3}[2]{*}{Mistral 7B} & vanilla & 52.3  & 63.5  & 71.6  & 58.6  & 46.0  \\
          & LoRA  & 62.3  & 70.0  & 84.2  & 96.2  & 48.8  \\
          & tanh  & \textbf{66.3 } & \textbf{72.4 } & \textbf{86.7 } & \textbf{96.8 } & \textbf{51.3 } \\
    \bottomrule
    \end{tabular}%
    }
  \caption{The performance for Qwen2.5-7B ($\xi=5$) and Mistral-7B ($\xi=3$) with our activation function}
  \label{tab:per_qwen_mis}%
\end{table}%

\subsection{Implementation Details}
When calculating $\Delta \mW$, we use Low Rank Adaption, which means we actually calculates $\Delta \mW=\mA \cdot \mB$, $\mA \in \mathbb{R}^{h \times r}$, where $h$ is the hidden dimension and $r$ is the rank, in our experiments, we set $r=64$. For the experiments, if not otherwise specified, we position the query before the documents.


When calculating the attention, we use Group Query Attention like used in the LLaMA architecture. And we use Low Rank Adaption with rank 64, so the parameters for training are the same. We conduct our experiments on Llama-3.1-8B-Instruct, Qwen/Qwen2.5-7B-Instruct, mistralai/Mistral-7B-Instruct-v0.2.

It is worth noting that, our method does not need to calculate another attention matrix and add the new attention matrix to the previous one. We can directly add the activation to the existing attention matrix. However, directly add the activation would greatly disrupt the attention patter, causing bad performance, therefore, we set $\xi$ to be $0$ at first so our method becomes vanilla attention, then during the training process, we gradually increase $\xi$ linearly for the first 80\% steps and keeps $\xi$ a constant for the last 20\% steps. And besides LoRA fine-tuning, We conduct full fine-tuning under this setting and the result is shown in Table \ref{tab:full_finetuning}.

We compare our method with LoRA mainly because our method focus on how to adjust the attention schema to make the model more robust to noise. But current researchs primarily focuses on either how to structure training data \citep{RAG_robust,RAG_robust_finetune_25} or how to train models to better handle different types of noise \citep{RAAT}. In contrast, our work addresses a different challenge: the standard self-attention mechanism inherently struggles to filter out noisy information So we modify the attention computation by incorporating a non-linear activation function to enhance robustness.
Differential Transformer \citep{differentialtransformer} also tries to adjust attention, but requires to train the model from scratch instead of fine-tuning based on existing model, and CrAM does not involve fine-tuning, so the comparison is unfair.

\subsection{Main Results}

\begin{figure*}
    \centering
    \includegraphics[width= \linewidth]{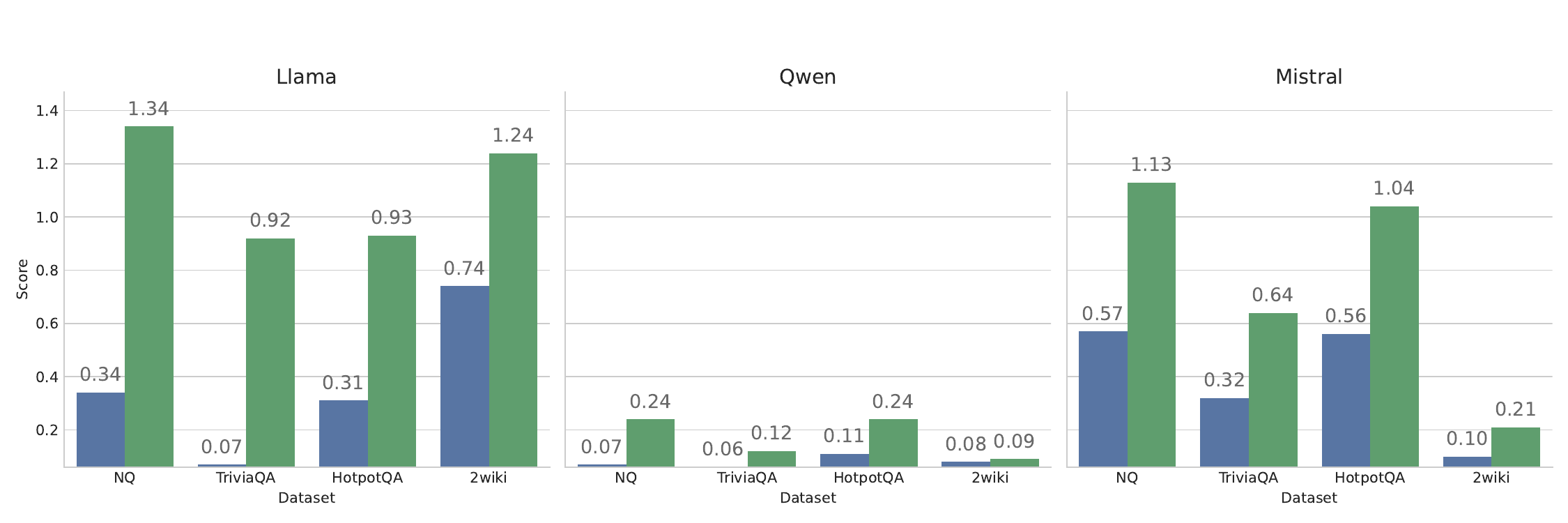}
    \caption{The difference of attention score on answer tokens (mean(attn(answer))-mean(attn(other)), for clarity, we scale this gap by a factor of 1,000.}
    \label{fig:margin_atten}
\end{figure*}

Table \ref{tab:per_noise} shows the performance when faced with explicit distracting documents, we evaluate two different setting where the query is placed behind the documents or ahead the documents. The result shows that with $g(x)$, the model can better distinguish between relevant and distracting documents, showing better performance. Also, we can observe that after fine-tuning, placing the query ahead of documents still helps. The result shown in Table \ref{tab:per_noise} might be high especially for 2wiki, this is mainly because we add gold documents to the context to show how can the model grab useful information from context. We also show the performance when all documents are retrieved in Table \ref{tab:all_retrieval_per}, it also shows that our method performs well.

\begin{figure}
    \centering
    \includegraphics[width=\linewidth]{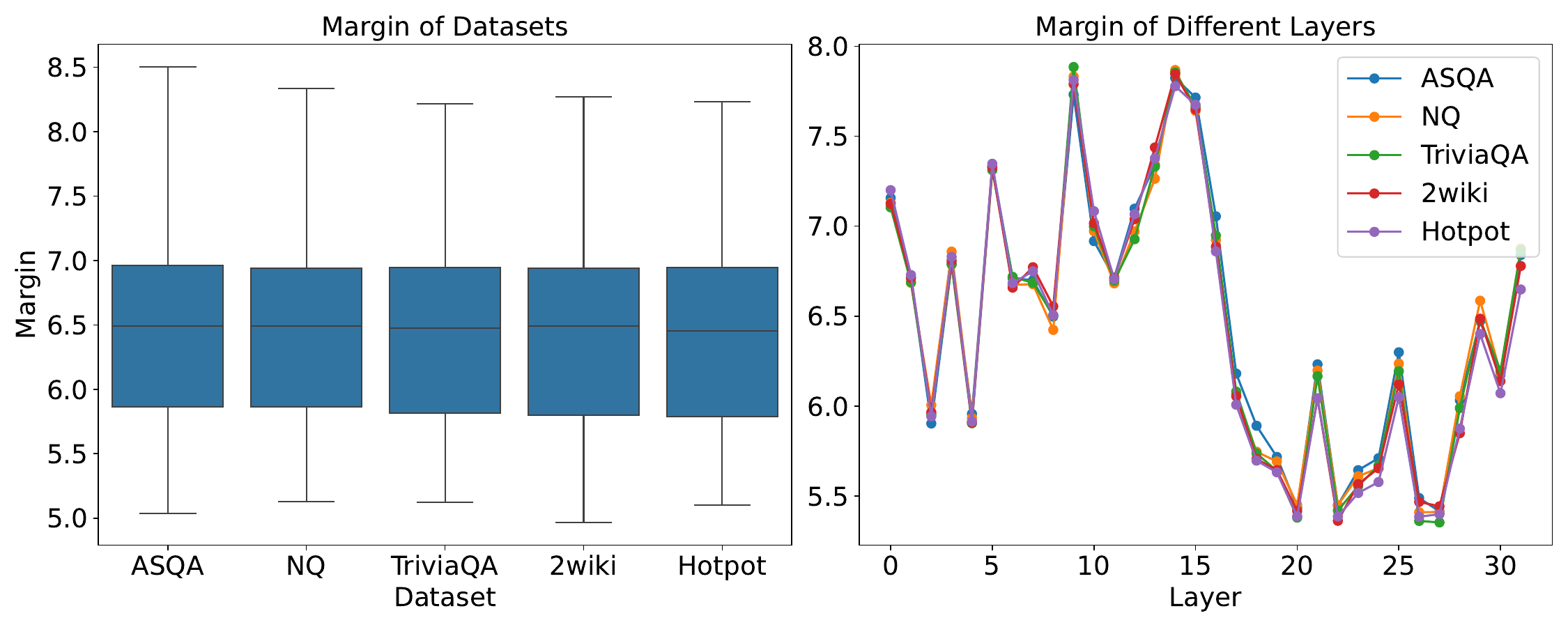}
    \caption{The margin of attention scores. We calculate the margin as the difference between the 90th and 10th percentile attention scores to reduce the influence of outliers}
    \label{fig:margin}
\end{figure}

And we can observe that the hyperparameter $\xi$ actually does not require specific tuning, we can directly set the value based on the inherent attention weight margin of the model, which means the gap between high attention scores and low attention scores. If we set $\xi$ to make it cover the margin of attention scores, then our method can work well.
We show the margin of the original attention weight of Llama 3.1-8B-Instruct in Figure \ref{fig:margin}, we can observe that most of the margins of Llama lies about 6, so with $\xi=3$, (margin between (-3,3)), the model can effectively distinguish irrelevant tokens, and larger values like $\xi=5$ or $\xi=10$ also performs good as we show in Table \ref{tab:per_noise}.

We also conduct experiments on Qwen/Qwen2.5-7B-Instruct and mistralai/Mistral-7B-Instruct-v0.2, we show the performance in Table \ref{tab:per_qwen_mis}, which shows that our method also effectively helps the performance on Qwen2.5-7B-Instruct and Mistral-7B-Instruct. We also show the performance with full fine-tuning in Table \ref{tab:full_finetuning}.

\subsection{The Difference of Attention Score}

By adding the activation function, our method can effectively distinguish between useful information and noise, to show this, we calculate the attention score gap between the tokens containing answer and other tokens (after softmax, mean(attn(answer))-mean(attn(other)), for clarity, we scale this gap by a factor of 1,000, the result is show in Figure \ref{fig:margin_atten}.
We do not show the result of ASQA because it is a long form QA, the answer is not directly indicated in the documents. The result shows that the rectification helps the model to recognize the answer and filter out noisy information, which explains why our method helps.

\subsection{Ablation Study with Different Activation}

\begin{figure}
    \centering
    \includegraphics[width= \linewidth]{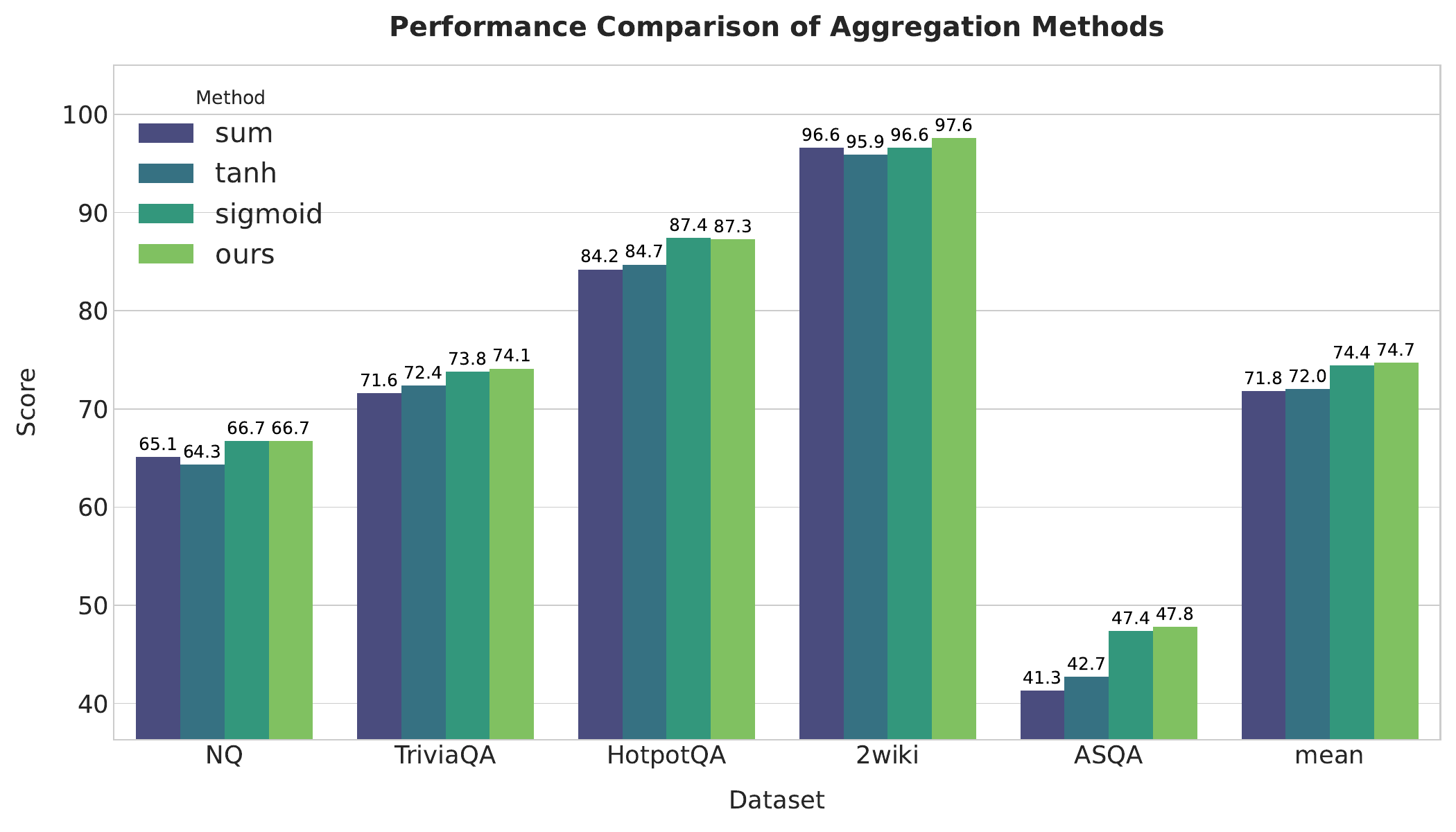}
    \caption{The performance when we set different activation function, $sum (\xi \cdot tanh(x),x)$ (sum), $\xi \cdot tanh(x)$ (tanh). We also show the performance when we use $g(x)=max/min(x,2 \xi \cdot (sigmoid(x)-0.5))$ (sigmoid). $\xi$ is set to 3, and ours means our method.}
    \label{fig:compare}
\end{figure}

We also conduct experiments when $g(x)=\xi \cdot tanh(x)$ and $g(x)=\xi \cdot tanh(x)+x$. $g(x)=tanh(x)$ stands for the situation that $g(x)$ only focus on enlarge the margin between relevant and irrelevant tokens without further linear growth. And $g(x)=tanh(x)+x$ stands for the situation where the the steady growth process is missing, it rapidly increase with $x$, so the soft clamp will not work. We also try using 
\begin{equation}
  g(x)=
  \begin{cases}
    \max(\xi \cdot (sigmoid(x)-0.5)), x) & \text{ if } \ x>=0, \\
    \min(\xi \cdot (sigmoid(x)-0.5)), x) & \text{ else, }
  \end{cases}
  \nonumber
\end{equation}
We use sigmoid activation instead of tanh and show the performance. As shown in Figure \ref{fig:compare}, simply use $tanh(x)$ or $tanh(x)+x$ has suboptimal performance, this is mainly because they fail to optimize the attention pattern on relevant tokens or missing the saturating behavior.
Also we can observe that replacing $tanh$ with $sigmoid$ helps the performance, this is mainly because $sigmoid$ is actually quite similar with $tanh$, they all increases fast at the beginning and result in a steady region after the growth.

\section{Conclusion}

In this paper, we highlight that noise filtering in RAG is inherently difficult, limited number of transformer layers can not effectively solve it, so we require the LLM to be robust to noise information. Then we show that simply fine-tuning the LLM may not be optimal as it will disturb the attention pattern. Then we propose a new fine-tuning method which can help to be more robust to noise, extensive experiments show that our method works well, it can effectively filter out noise while taking advantage of relevant information.

\section*{Limitations}
The paper discusses the limitation of LLM when dealing with noisy information, showing that current LLMs can not effectively process noisy information. However, although a new fine-tuning method is proposed, it can not fully address the problem as it is a fine-tuning based on the trained model. It might help more if we train a model from scratch, but due to limited computational resource, this can only leave for future work.

\bibliography{reference}

\appendix

\section{Experiments} \label{app:more_experiments}

\subsection{Experiment settings}
When conduct fine-tuning, we use learning rate of 1e-4, and we use kaiming initialization to initialize the parameter with $a=\sqrt{5}$. The experiments is conducted on 8 NVIDIA A100 80GB.

For NQ, TriviaQA, HotpotQA and 2Wiki, we randomly select 3000 samples to test and another 7000 to train. For ASQA we use the split of ALCE \citep{ALCE} and use the 948 samples to test the performance and another 4000 for training. Also, we train 3 epochs for each dataset except ASQA due to its limited number of data, we train it for 5 epochs instead. We train the model with batch size 8.

We use DPR as the retriever and retrieve top 3 noise documents as noise. Then we mix it with the gold documents and shuffle the documents randomly as the input context.

\subsection{Placing the Query Ahead Helps}

Here we show  that placing the query ahead of the documents can help the performance, we conduct experiments on  NQ, TriviaQA, HotpotQA, 2Wiki and ASQA. we conduct evaluation on 3000 samples for the first 4 datasets and 948 for ASQA. The result shown in Table \ref{tab:reverse_performance} shows that placing the query ahead can indeed help the performance.

\begin{table*}[htbp]
  \centering
\resizebox{\textwidth}{!}{%
    \begin{tabular}{ccccccccccc}
    \toprule
          & \multicolumn{2}{c}{NQ} & \multicolumn{2}{c}{TriviaQA} & \multicolumn{2}{c}{HotpotQA} & \multicolumn{2}{c}{2wiki} & \multicolumn{2}{c}{ASQA} \\
          & reverse & vanilla & reverse & vanilla & reverse & vanilla & reverse & vanilla & reverse & vanilla \\
    \midrule
    GPT-4o & \textbf{61.24} & 56.31 & \textbf{72.35} & 70.61 & \textbf{82.41} & 80.34 & \textbf{85.73} & 83.15 & \textbf{49.16} & 45.13 \\
    DeepSeek & \textbf{59.69 } & 56.91  & \textbf{70.57 } & 67.17  & \textbf{79.48 } & 76.32  & \textbf{80.80 } & 72.71  & \textbf{47.99 } & 45.62  \\
    Llama 80B & \textbf{63.13 } & 54.07  & \textbf{71.07 } & 53.93  & \textbf{81.53 } & 76.83  & \textbf{87.13 } & 75.80  & \textbf{48.58 } & 48.33  \\
    Llama 8B & \textbf{52.43 } & 46.70  & \textbf{52.07 } & 45.63  & \textbf{61.43 } & 54.33  & \textbf{53.70 } & 52.07  & \textbf{42.04 } & 41.81  \\
    Mistral 7B & \textbf{52.27 } & 51.63  & \textbf{63.50 } & 63.50  & \textbf{71.57 } & 70.10  & \textbf{58.57 } & 55.30  & \textbf{46.04 } & 43.67  \\
    Qwen2.5 7B & 53.47  & \textbf{53.63 } & \textbf{59.87 } & 56.57  & \textbf{74.80 } & 70.47  & \textbf{67.07 } & 59.70  & 43.47  & \textbf{44.74 } \\
    \bottomrule
    \end{tabular}%
    }
  \caption{The performance of 3000 samples except ASQA (948 samples), this shows that putting the query ahead could help the performance.}
  \label{tab:reverse_performance}%
\end{table*}%

\begin{table*}[htbp]
  \centering

    \begin{tabular}{cccccccc}
    \toprule
          &       & NQ    & TriviaQA & HotpotQA & 2wiki & ASQA  & mean \\
    \midrule
    \multirow{3}[2]{*}{Llama} & vanilla & 21.8  & 33.7  & 14.7  & 28.8  & 24.5  & 24.7 \\
          & LoRA  & 37.2  & 52.3  & 32.5  & 51.2  & 30.4  & 40.72 \\
          & ours  & \textbf{38.4} & \textbf{56.2} & \textbf{34.7} & \textbf{52.4} & \textbf{32.7} & \textbf{42.88} \\
    \midrule
    \multirow{3}[2]{*}{Qwen} & vanilla & 20.3  & 29.2  & 13.8  & 28.3  & 24.2  & 23.16 \\
          & LoRA  & 26.2  & 45.5  & 27.2  & 49.8  & 34.2  & 36.58 \\
          & ours  & \textbf{28.1} & \textbf{49.4} & \textbf{30.1} & \textbf{52.4} & \textbf{35.7} & \textbf{39.14} \\
    \midrule
    \multirow{3}[2]{*}{Mistral} & vanilla & 26.4  & 37.8  & 24.9  & 43.5  & 24.8  & 31.48 \\
          & LoRA  & 40.7  & 56.1  & \textbf{37.7} & 52.8  & 33.2  & 44.1 \\
          & ours  & \textbf{42.5} & \textbf{58.3} & 37.2  & \textbf{55.3} & \textbf{35.2} & \textbf{45.7} \\
    \bottomrule
    \end{tabular}%
    \caption{The performance when all documents are retrieved. We retrieve top 5 documents and use those retrieved documents as context. Our method also shows better performance}\label{tab:all_retrieval_per}%
\end{table*}%

\begin{table*}[htbp]
  \centering

    \begin{tabular}{cccccccc}
    \toprule
          &       & NQ    & TriviaQA & HotpotQA & 2wiki & ASQA  & mean \\
    \midrule
    \multirow{2}[2]{*}{Llama} & SFT   & 67.8  & 74.6  & 88.7  & 96.4  & 45.7  & 74.6  \\
          & ours  & 69.2  & 76.4  & 90.2  & 96.7  & 49.3  & 76.4  \\
    \midrule
    \multirow{2}[2]{*}{Qwen} & SFT   & 65.2  & 72.3  & 85.1  & 96.1  & 49.2  & 73.6  \\
          & ours  & 67.5  & 75.1  & 87.3  & 96.1  & 52.1  & 75.6  \\
    \midrule
    \multirow{2}[2]{*}{Mistral} & SFT   & 65.2  & 73.6  & 86.1  & 96.4  & 50.3  & 74.3  \\
          & ours  & 67.6  & 75.2  & 88.3  & 97.8  & 53.6  & 76.5  \\
    \bottomrule
    \end{tabular}%
  \caption{The performance of full fine-tuning, and ours means adding the rectification directly to the existing attention matrix, which means we only need to calculate one attention score.}
  \label{tab:full_finetuning}%
\end{table*}%

\begin{figure}[htbp]
    \centering
    \begin{subfigure}{0.48\textwidth}
      \centering
      \includegraphics[width=\linewidth]{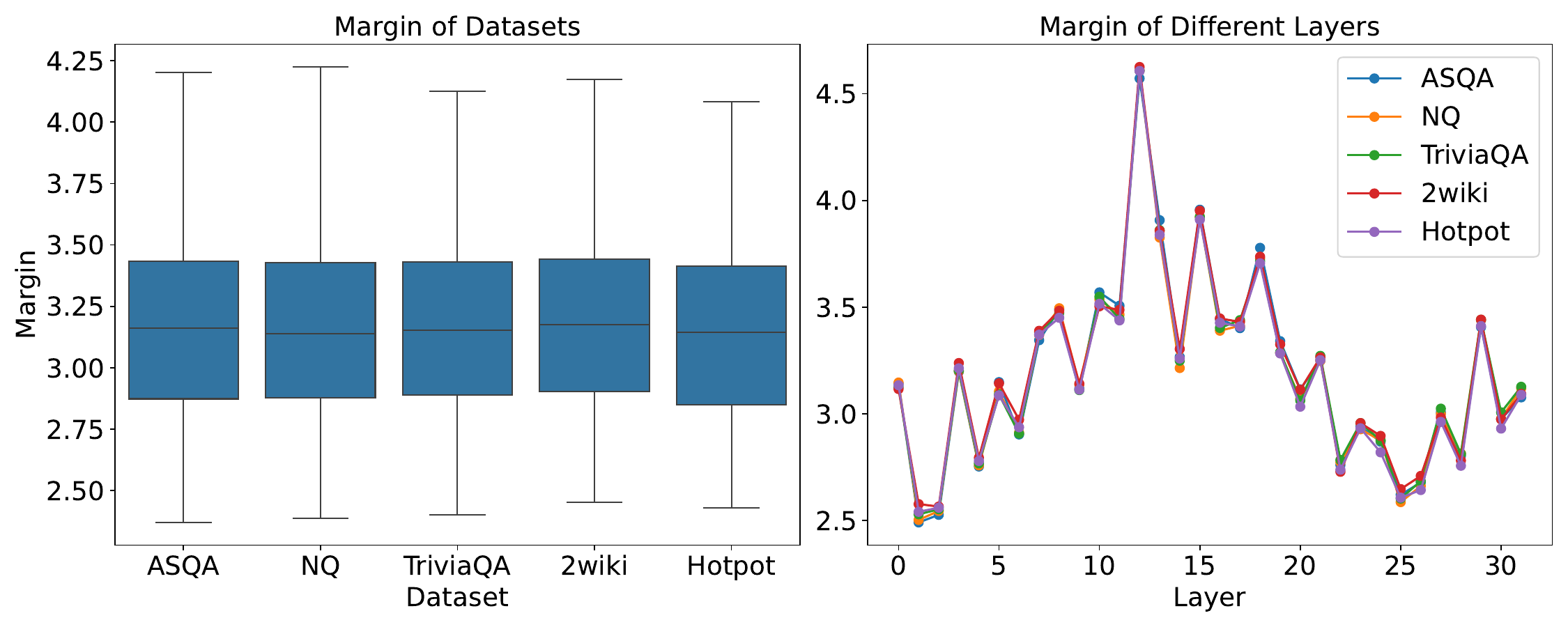}
      \caption{The margin of attention scores for Mistral 7B}
      \label{fig:margin_mis}
    \end{subfigure}
    \hfill
    \begin{subfigure}{0.48\textwidth}
      \centering
      \includegraphics[width=\linewidth]{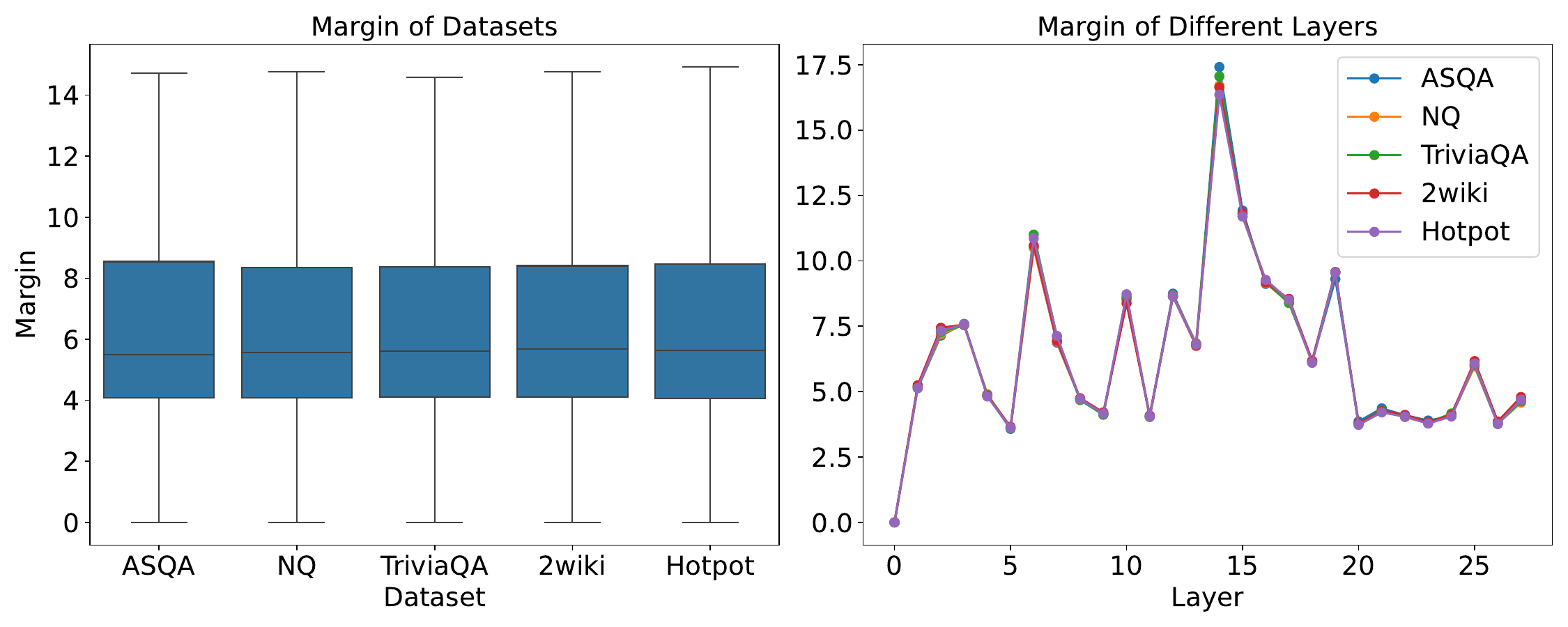}
      \caption{The margin of attention scores for Qwen 7B}
      \label{fig:margin_qwen}
    \end{subfigure}
    \caption{The margin of attention scores. We calculate the margin as the difference between the 90th and 10th percentile attention scores to reduce the influence of outliers}
    \label{fig:combined}
\end{figure}

\onecolumn
\section{Proofs}

\subsection{The Triple-Wise problem} \label{proof_3match}

  Suppose Alice and Bob are given inputs $a, b \in \{0,1 \}^n$, respectively, with the goal of jointly computing $\mathrm{DISJ}(a, b) = \max_i a_i b_i$ by alternately sending a single bit message to the other party over a sequence of communication rounds.
  Any deterministic protocol for computing $\mathrm{DISJ}(a, b)$ requires at least $n$ rounds of communication.

\begin{equation}
  r_i=\begin{cases}
    0 & \text{ if }\ \exists \ a,b \ s.t. \ g(x_i,x_a,x_b)=0 \\
     1 & \text{ else }
   \end{cases}
   \nonumber
\end{equation}

In normal cases, judging the value of $r_i$ requires calculating $g(x_i,x_a,x_b)$ for all $a \in [0,n_d)$ and $b \in [n_d,n_d+n_q)$. Here we simplify the question, and we consider the situation where $g(x_i,x_a,x_b)=0$ only if $b=a+n_d$ and $n_q=n_d$. Apparently, this is a special case of the original problem, and if one layer of self-attention fail to solve this, it is impossible for it to solve the original problem.

If we assume that the input is like,
\begin{equation}\label{eq:domain}
  \vx_i \in 
  \begin{cases}\{\vx_i\} & \text{if} \ i = 0, \\ 
    \{0, \vx_a \} &\text{if} \ i \in \{1, \dots, n_d-1\}, \\ 
    \{0, \vx_b\} & \text{if} \ i \in \{n_d, \dots, 2 \cdot n_d-1\}.
  \end{cases}
\end{equation}
Given input $(a,b) \in \{0,1 \}^{n_d} \times \{0,1 \}^{n_d} $, let $x_i=x_a$ if and only if $a_i=1$ and let $x_i=x_b$ if and only if $b_{i-n_d}=1$. In this way $r_i=0$ if and only if $\mathrm{DISJ}(a,b)=1$.

For simplicity, we use $n=n_d$. If we consider the setting of RAG, then actually Alice and Bob each hold a matrix $\mA \in \mathbb{R}^{n\times d}$, $\mB \in \mathbb{R}^{n\times d}$,
each row of the matrix contains $\vw$, which is the embedding to judge if the token a relevant information. So $d=H(\vw)$. and we assume that $\vx=[\vs,\vw]$.

Then, to judge is a token relevant, we need to embedding $x_i$ to contain information about $\vw$, if we want to judge the relevance of $x_0$, then

\begin{equation}
  \vx_i = 
  \begin{cases} \vx_i & \text{if} \ i = 0, \\ 
    [\vs, \va_i] &\text{if} \ i \in \{1, \dots, n_d-1\}, \\ 
    [\vs, \vb_i] & \text{if} \ i \in \{n_d, \dots, 2 \cdot n_d-1\}.
  \end{cases}
\end{equation}

Let $\mathrm{DISJ1}(\mA,\mB)=\max_i (g'(\vx_i,\va_i,\vb_i))$ to be noise filtering task in RAG, then it requires to access $\vw$ of all tokens, which means the calculation of $DISJ1$ requires $n \times H(\vw)$ bits of communication.

Also similar to the setting of $x$, $r_i=1$ if and only if  $\mathrm{DISJ1}(\mA,\mB)=1$.

Then, this is the same with the 3Match problem, following the proof of Theorem 7 in \citet{3Match}, and with the following form of transformer $f(\mX)$, $2pH \log \log n+mpH \log \log n \approx mpH \log \log n$ bits are communicated.

$$f(\mX)=\phi(f_h(\mX)),$$ where $\phi$ stands for the feed forward layers.

$$f_h(\mX)=\frac{\sum_{i=1}^N \exp \left((\mW_q x_1)^T \mW_k x_i \right)\mW_v x_i }{\sum_{i=1}^N  \exp \left((\mW_q x_1)^T \mW_k x_i\right)}$$

Therefore, only we require $mpH \log \log n \geq n H(\vw) \rightarrow mph \geq n H(\vw)/ \log \log n$.

\begin{theorem}
  For input documents of length $n$, if $mpH \leq \Omega (n H(\vw)/ \log \log n)$, then there is no one layer transformer $\gM$ with embedding size $m$, precision $p$ and $H$ heads satisfying $\gM(X) = r$.
\end{theorem}
Also, as shown in \citet{3Match}, multiple layers of multi-headed
attention are subject to the same impossibility

\begin{conjecture}
  Every multi-layer transformer that computes Match3 must have width, depth, embedding dimension, or bit complexity at least $N^{\Omega(1)}$.
\end{conjecture}

This is based on the situation that, each translation only need 1 bit, for noise filtering, we require to translate $H(\vw)$ bit of information each time, so it requires width, depth, embedding dimension, or bit complexity at least $N^{\Omega(1)} \cdot H(\vw)$. This directly means that triple-wise problems can not be solved with limited number of transformer layers

\subsection{Proof of Theorem \ref{theorem:lora_fail}} \label{proof_lora_fail}

Let $attn(x_i)=(\mW_q \vx_i)^T \mW_k \mX_{:i}$ be the original attention layer of LLM, and $attn'(x_i)=\left((\mW_q+\Delta \mW_q) \vx_i\right)^T  (\mW_k+\Delta \mW_k)\mX_{:i}$ be the fine-tuned one and $\widehat{attn}$ be desired function, can we fine-tune the model to be $\widehat{attn}$?

So we need 

\begin{equation}
  softmax(attn'(\mX))[i] \approx  
  \begin{cases}
     0 & \text{ if } \ \exists \ b \ s.t. \ g_1(x_i,x_b)=0 \\
     softmax(attn(\mX_r))[i] & \text{ else }
   \end{cases}
   \nonumber
\end{equation}
where $\mX_r$ means those related tokens. let $\mA_{i,j}=attn(\vx_i,\vx_j)=(\mW_q \vx_i)^T \mW_k \vx_j=\vx_i^T \mW \vx_j$

\begin{equation}
  \begin{split}
    attn'(\vx_i,\vx_j)
    &=\left((\mW_q+\Delta \mW_q)  \vx_i \right)^T(\mW_k+\Delta \mW_k)\vx_j\\
    &=\vx_i (\mW+\Delta \mW) \vx_j\\
    &=\vx_i \mW \vx_j+\vx_i \Delta \mW \vx_j
  \end{split}
   \nonumber
\end{equation}
where $\Delta \mW=\Delta \mW_q \mW_k+\mW_q \Delta \mW_k+\Delta \mW_q \Delta \mW_k$

Then, to effectively separate noise information, we require the attention score of the noise to be small and the attention score of useful information to be large, so to filter out noise, we require $\Delta \mW_q', \Delta \mW_k'$ satisfying 
\begin{equation}
  \vx_i \Delta \mW \vx_j
  \begin{cases}
    \leq c_l & \text{if $x_j$ is noise}, \\
    \in [c_h,c_h+\xi_{r}]& \text{else},
  \end{cases}
\end{equation}
where $c_l$ and $c_h$ are constants and $c_h>c_l$

if $\vx_j$ is a noise token and $\vx_i \Delta \mW \vx_j=c_l$, then
\begin{equation}
  \begin{split}
    softmax(attn'(\vx_i))[j]-0
    &=softmax(A_{i,:}+x_i \Delta \mW \mX)[j]\\
    &=\frac{\exp(A_{i,j}+x_i \Delta \mW x_j)}{\sum_k \exp(A_{i,k}+\vx_i \Delta \mW \vx_k)}\\
    &=\frac{\exp(A_{i,j}+c_l)}{\sum_k \exp(A_{i,k}+\vx_i \Delta \mW \vx_k)}\\
    &=\frac{\exp(A_{i,j}+c_l-c_h)}{\sum_k \exp(A_{i,k}+\vx_i \Delta \mW \vx_k -c_h)}\\
  \end{split}
   \nonumber
\end{equation}
if we need $softmax(attn'(\vx_i))[j]-0 \leq \epsilon$, let $A_{i,j}=\max(A_{i,:})$, then 
\begin{gather*}
  \frac{\exp(A_{i,j}+c_l-c_h)}{\sum_k \exp(A_{i,k}+\vx_i \Delta \mW \vx_k -c_h)}\leq softmax(attn'(\vx_i))[j]-0 \leq \epsilon\\
  \exp(A_{i,j}+c_l-c_h) \leq \epsilon \sum_k \exp(A_{i,k}+\vx_i \Delta \mW \vx_k -c_h)\\
  c_l-c_h \leq \ln \left( \epsilon \sum_k \exp(A_{i,k}+\vx_i \Delta \mW \vx_k -c_h)\right)-A_{i,j}\\
  c_l-c_h \leq \ln \left( \epsilon n \exp(A_{i,j}+\xi_{r}) \right)-A_{i,j}\\
  c_l-c_h \leq \xi_{r} +\ln \epsilon n
\end{gather*}

else if $\vx_j$ is a relevant token, let $c=c_h$, and $\sum_{k'} \exp(A_{i,k'})$ denotes the summation of all relevant tokens, we consider a simple case where $\vx_i \Delta \mW \vx_j=c_h$, and for one token $\vx_{k1}$, we have $\vx_i \Delta \mW \vx_{k1}=c_h+\xi_{r}$, and for all other relevant tokens we have $\vx_i \Delta \mW \vx_{k}=c_h$

\begin{equation}
  \begin{split}
    &\hspace*{1.5em} softmax(\hat{attn}(\vx_i))[j]-softmax(attn'(\vx_i))[j]\\
    &=\frac{\exp(A_{i,j})}{\sum_{k'} \exp(A_{i,k'})}-\frac{\exp(A_{i,j}+x_i \Delta \mW x_j)}{\sum_{k} \exp(A_{i,k}+x_i \Delta \mW x_{k})}\\
    &=\frac{\exp(A_{i,j})}{\sum_{k'} \exp(A_{i,k'})}-\frac{\exp(A_{i,j}+x_i \Delta \mW x_j-c)}{\sum_{k} \exp(A_{i,k}+x_i \Delta \mW x_{k}-c)}\\
    &= \frac{\exp(A_{i,j})}{\sum_{k'} \exp(A_{i,k'})}-\frac{\exp(A_{i,j})}{\sum_{k} \exp(A_{i,k}+x_i \Delta \mW x_{k}-c)}\\
  \end{split}
\end{equation}
considering $softmax(\hat{attn}(\vx_i))[j]-softmax(attn'(\vx_i))[j]=c(\frac{1}{a}-\frac{1}{b})\leq \epsilon \frac{c}{a}$, $c=\exp(A_{i,j})$, $a=\sum_{k'} \exp(A_{i,k'})$, $b=\sum_{k} \exp(A_{i,k}+x_i \Delta \mW x_{k}-c)$
\begin{gather*}
  b-a \leq \epsilon b\\
  (1-\epsilon) b \leq a\\
  b \leq \frac{a}{1-\epsilon}\\
  \sum_{k} \exp(A_{i,k}+x_i \Delta \mW x_{k}-c) \leq \frac{\sum_{k'} \exp(A_{i,k'})}{1-\epsilon}\\
  \sum_{k'} \exp(A_{i,k'}+x_i \Delta \mW x_{k}-c) \leq \frac{\sum_{k'} \exp(A_{i,k'})}{1-\epsilon}\\
  \sum_{k'-k1} \exp(A_{i,k'})+\exp(A_{i,k1}+\xi_{r}) \leq \frac{\sum_{k'} \exp(A_{i,k'})}{1-\epsilon}\\
  \exp(A_{i,k1}+\xi_{r}) \leq \frac{\epsilon}{1-\epsilon} \sum_{k'-k} \exp(A_{i,k'})+\frac{\exp(A_{i,k1})}{1-\epsilon}\\
  \xi_{r} \leq \ln \left(\frac{\epsilon}{1-\epsilon} \sum_{k'-k} \exp(A_{i,k'})+\frac{\exp(A_{i,k1})}{1-\epsilon} \right)-A_{i,k1}
\end{gather*}

Considering the case that $\frac{\epsilon}{1-\epsilon} \sum_{k'-k} \exp(A_{i,k'}) \approx 0$, then we need $\xi_{r} \lesssim \ln \frac{1}{1-\epsilon}$

As $\epsilon \approx 0$, so $\ln \frac{1}{1-\epsilon} \approx 0$

\subsection{MLP also fails to filter out noise} \label{proof_mlp_fail}
In the following, we make use of the quantities
\begin{equation}
N_{max}(t) = \max_{\hat{v} \in \hat{\mathcal{V}}} N_{\hat{v}}(t), \qquad N_{min}(t) = \min_{\hat{v} \in \hat{\mathcal{V}}} N_{\hat{v}}(t),
\nonumber
\end{equation}
where
\begin{equation}
N_{\hat{v}}(t) = \sum_{v \in \mathcal{V}} \mathbb{1} \{ d(v,\hat{v}) \le t \}
\nonumber
\end{equation}
counts the number of $v \in \mathcal{V}$ within a ``distance'' $t$ of $\hat{v} \in \hat{\mathcal{V}}$. 

\begin{theorem} \label{thm:Partial}
  {\em (Fano's inequality with approximate recovery in \citet{fano_introductory})}
  For any random variables $v,\hat{v}$ on the finite alphabets $\mathcal{V},\hat{\mathcal{V}}$, we have
  \begin{equation}
  P_e(t) \ge \frac{H(\vv|\hat{\vv}) -1}{ \log\frac{|\mathcal{\mathcal{V}}|}{N_{max}(t)} }. \label{eq:Partial3}
  \end{equation}
  $P_e(t)=\Pr(||\vz-\hat{\vz}||>t)$, where $\vz$ is inferenced by some function and the input is $\vv$.
\end{theorem}

Assume the input to the feed forward layer $\vv$, the first $l$ layers are used to identify the relevance and the last few layers are used for inference. The output of the first $l$ layers are $\vv_l$, and the embedding is used to conduct inference and get the result $\vz$. So we can say that with probability $p \geq \frac{H(\vv_l|\hat{\vv}) -1}{ \log\frac{|\mathcal{\mathcal{S}}|}{N_{max}(t)} }$ the resulting embedding fails to be $t$ close to the original one \ie $d(\vz,\hat{\vz})\leq t$, where $\hat{\vv}$ stands for the optimal input  where all irrelevant information is filtered and all the related information is contained and $\hat{\vz}$ is the corresponding output..

Assume that $\hat{\vw}$ are equally distributed to each token which satisfies $I(\vw_i;\vv)=I(\vw_j;\vv)$, therefore, for each document, the model holds the same probability of mistakenly identify its relevance. Let $p_{we}$ stands for the error probability of identify noise tokens, and $\delta$ be the percentage of relevant tokens. With probability $\delta p_{we}$, the relevant token is mistakenly regarded as irrelevant, and with probability $(1-\delta)p_{we}$ the irrelevant token is mistakenly regarded as relevant.
So there are $\frac{\delta (1-p_{we})}{\delta (1-p_{we})+(1-\delta)p_{we}}$ percent of information about the relevant ones. Also $p_{we}$ percent of relevant information and $1-p_{we}$ percent of irrelevant information are discarded, then $I(\vs;\vv_l)=\left((1-p_{we}) \cdot \delta +p_{we} \cdot (1-\delta) \right)I(\vs;\vv)$ are the left information about inference, among these, $\frac{\delta (1-p_{we})}{\delta (1-p_{we})+(1-\delta)p_{we}}$ are acutally related information, others are noisy information.

In this way,
\begin{equation}
  \begin{split}
    I(\vv_l;\hat{\vv})&=I(\vv_l;\vs)\\
    &=\frac{\delta (1-p_{we})}{\delta (1-p_{we}) +(1-\delta)p_{we}} \cdot \left((1-p_{we}) \cdot \delta +p_{we} \cdot (1-\delta) \right)I(\vs;\vv)\\
    &=\delta (1-p_{we}) \cdot I(\vs;\vv)\\
    &\leq \delta (1-\frac{H(\vw|\vv-1)}{H(\vw)}) \cdot I(\vs;\vv)\\
    &=\delta (\frac{I(\vw;\vv)+1}{H(\vw)})\cdot I(\vs;\vv)
  \end{split}
  \nonumber
\end{equation}


Therefore,
\begin{equation}
  \begin{split}
    P_e(t)
    &\ge \frac{H(\vv|\hat{\vv}) -1}{ \log\frac{|\mathcal{\mathcal{V}}|}{N_{max}(t)} }=\frac{H(\vv)-I(\vv;\hat{\vv})}{\log\frac{|\mathcal{\mathcal{V}}|}{N_{max}(t)}}\\
    &\geq \frac{H(\vv)-g_1(\delta,I(\vw;\vv)) \cdot I(\vs;\vv)}{\log\frac{|\mathcal{\mathcal{V}}|}{N_{max}(t)}}
  \end{split}
  \nonumber
\end{equation}
where $g_1(\delta,I(\vw;\vv))=\delta (\frac{I(\vw;\vv)+1}{H(\vw)})$

So when there is no noise, the inference can be conducted based on those information, then we have 
\begin{equation}
  \begin{split}
    \Pr \left(||z-\hat{z}|| > t\right) &\geq \frac{H(\vv)-g_1(\delta,I(\vw;\vv)) \cdot I(\vs;\vv)}{\log\frac{|\mathcal{\mathcal{V}}|}{N_{max}(t)}}\\
    \Pr \left(||z-\hat{z}|| \leq t\right) &\leq 1-\frac{H(\vv)-g_1(\delta,I(\vw;\vv)) \cdot I(\vs;\vv)}{\log\frac{|\mathcal{\mathcal{V}}|}{N_{max}(t)}}
  \end{split}
  \nonumber
\end{equation}
Considering the noise in the embedding of $\vv_l$, and the noise would have negative impact on the inference.

Also the extra noisy information contained in $\vv_l$ is 
\begin{equation}
  \begin{split}
    I(\vv^-;\vv_l)&=\frac{(1-\delta)p_{we}}{\delta (1-p_{we})+(1-\delta)p_{we}} \cdot  \left((1-p_{we}) \cdot \delta +p_{we} \cdot (1-\delta) \right)I(\vs;\vv)\\
    &=(1-\delta)p_{we} \cdot I(\vs;\vv)\\
    &\geq(1-\delta) \frac{H(\vw|\vv)-1}{H(\vw)}\cdot I(\vs;\vv)
  \end{split}
  \nonumber
\end{equation}

Consider the best case where $I(\vv^-;\vv_l)=(1-\delta) \frac{H(\vw|\vv)-1}{H(\vw)}\cdot I(\vs;\vv)$

\begin{theorem}[Theorem 2 of \citet{IB_generalization}] \label{thm:2}
  Let $\gD \subseteq \{1,2,\dots,D+1\}$.    Then,  for any $\delta>0$, with probability at least $1-\delta$ over the training set $s$, the following generalization bound holds: 
  \begin{align}  \label{eq:11} 
  \Delta(s)
  \le \min_{l \in \gD} Q_l,
  \end{align}
  where for $l \le D$,
  $$
  \scalebox{1.2}{%
  \begin{small}
  $
  Q_l \hspace{-2pt} = G_3^l \sqrt{\frac{\left( I_{}(X;Z_{l}^s|Y)+I(\phi_{l}^{S}; S) \right) \ln(2)+ \widehat \gG_2^l}{n}} + \frac{G_1^l(\zeta)}{\sqrt{n}};
  $
  \end{small}}
  $$ 
  and for $l = D+1,$
  $$Q_l=\gR(f^{s}) \sqrt{\frac{I_{}(\phi_{l}^{S}; S)\ln(2)  + \check \gG_2^l }{2n}},$$
  Here,   $S \sim\gP^{\otimes n}$,     $G_1^l(\zeta)=\hat{\gO}(\sqrt{I(\phi_{l}^{S}; S)+1})$,  $\widehat \gG_2^{l}=\hat{\gO}(1)$, $\check \gG_2^l=\hat{\gO}(1)$, and $G_3^{l}=\hat{\gO}(1)$ as $n\rightarrow \infty$. The formulas of $G_1^l(\zeta)$, $\widehat \gG_2^{l}$, $\check \gG_2^l$, and $G_3^l$  are given in Appendix.
\end{theorem}

using $||f(x)-\hat{f}(x)||$ as the loss function, then we have that
\begin{equation}
  \begin{split}
    \gL-\hat{\gL}
    &\leq G_3^l \sqrt{\frac{\left( I_{}(X;Z_{l}^s|Y)+I(\phi_{l}^{S}; S) \right) \ln(2)+ \widehat \gG_2^l}{n}} + \frac{G_1^l(\zeta)}{\sqrt{n}}\\
    &=c_1 \sqrt{I(\vv^-|\vv_l)+I(\phi_{l}^{S}; S)}+c_3\\
    &\leq c_1 \sqrt{I(\vv^-|\vv_l)}+c_1 \sqrt{I(\phi_{l}^{S}; S)}+c_3\\
    &\leq c_1 \sqrt{I(\vv^-|\vv_l)}+c_2
  \end{split}
\end{equation}
where $c_2=c_3+c_1 \sqrt{I(\phi_{l}^{S}; S)}$
Therefore,
\begin{equation}
  \Pr \left(||f(x)-\hat{f}(x)|| \leq t+c_1 \sqrt{I(\vv^-|\vv_l)+}+c_2 \right) \leq 1-\frac{H(\vv)-g_1(\delta,I(\vw;\vv)) \cdot I(\vs;\vv)}{\log\frac{|\mathcal{\mathcal{V}}|}{N_{max}(t)}}
\end{equation}
with $I(\vv^-;\vv_l)=(1-\delta) \frac{H(\vw|\vv)-1}{H(\vw)}\cdot I(\vs;\vv)=g_2(\delta,I(\vw;\vv)) \cdot I(\vs;\vv)$.

\begin{equation}
  \begin{split}
    &\Pr \left(||f(x)-\hat{f}(x)|| > t+c_1 \sqrt{g_2(\delta,I(\vw;\vv)) \cdot I(\vs;\vv)+}+c_2 \right) \\
    &\hspace{1.5em}  > \frac{H(\vv)-g_1(\delta,I(\vw;\vv)) \cdot I(\vs;\vv)}{\log\frac{|\mathcal{\mathcal{V}}|}{N_{max}(t)}}
  \end{split}
\end{equation}

\begin{theorem}
  For a Feed Forward Network $f$ and the input $x$ contains $1-\delta$ percent of noisy information, assume the optimal function is $\hat{f}(x)$ which filter out the noise and finish the inference, then
  \begin{equation}
    \begin{split}
      &\Pr \left(||f(x)-\hat{f}(x)|| > t' \right)   > \frac{H(\vv)-g_1(\delta,I(\vw;\vv)) \cdot I(\vs;\vv)}{\log\frac{|\mathcal{\mathcal{V}}|}{N_{max}(t)}},
    \end{split}
  \end{equation}
  where $t'=t+c_1 \sqrt{g_2(\delta,I(\vw;\vv)) \cdot I(\vs;\vv)}+c_2$. $g_1(\delta,I(\vw;\vv))=\delta (\frac{I(\vw;\vv)+1}{H(\vw)})$ $g_2(\delta,I(\vw;\vv))=(1-\delta) \frac{H(\vw|\vv)-1}{H(\vw)}$
\end{theorem}

\end{document}